\documentclass[preprint,12pt]{elsarticle}
\usepackage{graphicx}
\usepackage{amssymb}
\usepackage{lineno}
\usepackage{hyperref}
\usepackage{amsmath}
\usepackage{algorithm} 
\usepackage{algpseudocode} 
\usepackage{enumerate}
\journal{Computerized Medical Imaging and Graphics}
\usepackage{booktabs}
\usepackage{multirow} 
\usepackage{rotating}
\usepackage{pifont}
\usepackage{booktabs} 
\usepackage{tabularx} 
\usepackage{siunitx}  
\usepackage{ragged2e} 
\usepackage{array}    
\usepackage{subcaption}
\usepackage{tabularx} %
\usepackage{caption}  
\usepackage[table,xcdraw]{xcolor}
\newcolumntype{C}{>{\centering\arraybackslash}X}

\newcommand*{\updatedText}[1]{\textcolor{black}{ #1}}
\begin{document}

\begin{frontmatter}

\title{\updatedText{DAM-Seg: Anatomically accurate cardiac segmentation using Dense Associative Networks}}

\author[aff1]{Zahid~Ullah}
\author[aff1]{Jihie~Kim}

\address[aff1]{Division of AI Software Convergence, Dongguk University,
	Seoul 04620, Republic of Korea}

\begin{abstract}
Deep learning-based cardiac segmentation has seen significant advancements over the years. Many studies have tackled the challenge of anatomically incorrect segmentation predictions by introducing auxiliary modules. These modules either post-process segmentation outputs or enforce consistency between specific points to ensure anatomical correctness. However, such approaches often increase network complexity, require separate training for these modules, and may lack robustness in scenarios with poor visibility. To address these limitations, we propose a novel transformer-based architecture that leverages dense associative networks to learn and retain specific patterns inherent to cardiac inputs. Unlike traditional methods, our approach restricts the network to memorize a limited set of patterns. During forward propagation, a weighted sum of these patterns is used to enforce anatomical correctness in the output. Since these patterns are input-independent, the model demonstrates enhanced robustness, even in cases with poor visibility. The proposed pipeline was evaluated on two publicly available datasets, CAMUS and CardiacNet. Experimental results indicate that our model consistently outperforms baseline approaches across all metrics, highlighting its effectiveness and reliability for cardiac segmentation tasks.
\end{abstract}


\end{frontmatter}

\section{Introduction}
\label{intro}
With the advent of deep learning, cardiac image segmentation has become a cornerstone of cardiovascular research, playing a pivotal role in diagnostics, disease monitoring, treatment planning, and prognosis. Over the past decade, the field has undergone a significant transformation, fueled by advancements in deep learning methodologies. Non-invasive medical imaging modalities such as magnetic resonance imaging (MRI), computed tomography (CT), and ultrasound have provided critical support by offering detailed insights into cardiac structures. When combined with deep learning techniques, these imaging methods enable the development of efficient and accurate diagnostic tools for medical professionals.

Cardiac segmentation, a key application of deep learning, involves the semantic division of regions within cardiac images. Researchers have made strides in this area by refining existing segmentation methodologies and introducing novel approaches to achieve robust and precise results. For instance, Petitjean et al. \cite{RV_segmentation} conducted a comprehensive analysis of right ventricle segmentation techniques, benchmarking various methods against expert-annotated segmentation masks. Similarly, Tran et al. \cite{cardiac_fcn} employed a fully convolutional network (FCN) architecture \cite{fcn} to perform cardiac segmentation, demonstrating the efficacy of deep learning models in this domain.

Significant research efforts have been dedicated to enhancing the robustness of cardiac segmentation models across diverse datasets and distributions. For instance, Li et al. \cite{displacement_aware} introduced a displacement-aware shape encoding and decoding framework. In their approach, the input is compactly encoded while accounting for potential displacements and deformations. These encoded features are then decoded into multiple shapes, enabling the model to handle a broader range of shape variations. Additionally, they proposed a dynamically expanding mechanism that allows the model to adapt to increased complexity by incorporating new modules as needed, further improving robustness and scalability.

Cai et al. \cite{cross_domain} addressed cross-domain challenges by introducing a cross-domain mixup strategy. This method trains the cardiac segmentation model to generalize effectively on a target dataset by leveraging knowledge from a source dataset. By reducing the performance gap between the two domains, their approach enhances the model's generalization ability, ensuring consistent performance across varying datasets. These advancements collectively contribute to the development of more robust and adaptable cardiac segmentation frameworks.

Traditional deep learning techniques often struggle with cases where the cardiac structure is partially obscured, as the lack of visibility can hinder the formation of accurate boundaries. Researchers have addressed this challenge through innovative approaches. For example, Painchaud et al. \cite{VAE_wrapping} introduced a method that enhances segmentation performance by leveraging Variational Autoencoders (VAEs). Their architecture wraps implausible segmentation predictions toward valid cardiac shapes. In this approach, the VAE is trained independently to learn representations of anatomically correct cardiac structures, enabling the segmentation model to correct invalid outputs effectively.

Similarly, the Graph Convolutional Network (GCN) \cite{GCN} offers a unique solution to the problem of anatomically incorrect predictions. Instead of performing pixel-wise classification, GCN predicts key boundary points of the cardiac structure. It employs a graph neural network to model relationships between adjacent key points, effectively capturing spatial dependencies. By learning these relationships, the model ensures anatomically consistent predictions and maintains the structural integrity of the cardiac segmentation, even in challenging scenarios. These methods demonstrate significant progress in overcoming visibility-related limitations in cardiac segmentation tasks.

A quick and effective approach to maintaining anatomical accuracy in poorly visible cardiac images involves enabling the network to store a general representation of cardiac structures and leveraging this information to generate anatomically consistent segmentation masks. Ramsauer et al. \cite{SA_update} introduced a continuous dense associative memory unit, which employs a self-attention-based mechanism to dynamically update the stored memory states within the network. This concept, inspired by biological neural networks and their efficient pattern storage and retrieval capabilities, holds significant promise for cardiac segmentation.

In the context of cardiac imaging, such a memory unit can learn and store typical cardiac shapes and structures, providing a robust foundation for handling image quality variations and poor visibility. Integrating this mechanism can enhance the model's robustness, ensure consistent anatomical outputs, and improve performance, especially on small datasets. Building on these principles, we leverage the pattern-learning and storage capabilities of dense associative networks to enhance the segmentation performance of our proposed dense associative memory segmentation (DAM-Seg) model.

The code implementation of our method will be made publicly available upon publication to support reproducibility and further research. Code Link: \url{https://github.com/Zahid672/cardio-segmentation-main/tree/main/cardio-segmentation-main}.

The contributions of our research are as follows:
\begin{itemize}
    \item We propose learnable memory transformation matrices that convert the static memory space into query and value matrices, independent of the input, enabling a robust and flexible representation of stored patterns.
    \item We enhance the memory update mechanism to induce $m$ meta-stable states, focusing on generalizable patterns rather than memorizing all input variations, thereby improving robustness and efficiency.
    \item We integrate our memory module with the DPT architecture \cite{dpt}, leveraging its segmentation capabilities to achieve anatomically accurate and consistent cardiac segmentation.
\end{itemize}

\begin{table*}[!ht]
\centering
\scriptsize
\caption{Related studies and their methodologies with drawbacks.}
\color{black}
\label{table:research_methodology}
\scalebox{1.4}{
\begin{tabular}{|l|l|l|}
\hline
\textbf{Paper}                                         & \textbf{Methodology}                                                                                                                                                                                                                                                & \textbf{Drawbacks}                                                                                                                                                                                                         \\ \hline
Tran et al. \textbackslash{}cite\{cardiac\_fcn\}       & \begin{tabular}[c]{@{}l@{}}Uses FCN trained in an End-to\\ -End manner.\end{tabular}                                                                                                                                                                                & \begin{tabular}[c]{@{}l@{}}The segmentation of \\ cardiac structures, especially\\  at the boundaries, can be\\  challenging due to \\ fuzzy edge information.\end{tabular}                                                \\ \hline
Li et al. \textbackslash{}cite\{displacement\_aware\}  & \begin{tabular}[c]{@{}l@{}}Proposed a displacement-aware\\  shape encoding and decoding \\ mechanism where the input is \\ first encoded into a compact \\ form while keeping the \\ potential displacement or \\ deformations under \\ consideration.\end{tabular} & \begin{tabular}[c]{@{}l@{}}The method's ability to adapt\\  to multiple domains\\  might lead to overfitting on \\ the specific datasets \\ used, potentially limiting its \\ generalization to  unseen data.\end{tabular} \\ \hline
Cai et al. \textbackslash{}cite\{cross\_domain\}       & \begin{tabular}[c]{@{}l@{}}propose a cross-domain mixup  \\ mechanism that trains the cardiac\\  segmentation model to perform \\ well on a certain target dataset \\ using the information from the\\  source dataset.\end{tabular}                                & \begin{tabular}[c]{@{}l@{}}The method may struggle with\\  a very large\\  differences between modalities.\end{tabular}                                                                                                    \\ \hline
Painchaud et al. \textbackslash{}cite\{VAE\_wrapping\} & \begin{tabular}[c]{@{}l@{}}use variational Auto encoders to\\  improve the segmentation\\  performance of a pre-trained \\ segmentation model by wrapping\\  the implausible predictions towards \\ the valid cardiac shape.\end{tabular}                           & \begin{tabular}[c]{@{}l@{}}The need for millions of latent \\ vectors to be  stored in memory.\end{tabular}                                                                                                                \\ \hline
GCN \textbackslash{}cite\{GCN\}                        & \begin{tabular}[c]{@{}l@{}}Proposes a combination of a \\ convolutional encoder and a\\  graph decoder, it addresses \\ the problem of dealing with\\  anatomically incorrect \\ predictions.\end{tabular}                                                          & \begin{tabular}[c]{@{}l@{}}This kind of strict enforcement \\ of anatomical  rules might \\ potentially miss rare but valid \\ anatomical variations\end{tabular}                                                          \\ \hline
\end{tabular}}
 \end{table*}

\section{Background information}\label{related}
This section explores how associative memory models have evolved through the years and their applications, providing context for understanding dense associative memories. Table \ref{table:research_methodology} shows the relevant studies with drawbacks.

Associative memory research has its roots in early neural network models, with the Hopfield network \cite{hopfield_orig} being a seminal contribution. Introduced by John Hopfield in 1982, this model demonstrated how a network of interconnected neurons could store and retrieve patterns, mimicking the associative properties of human memory.

Krotov and Hopfield \cite{polynomial_DAM} generalize the energy function to $E = -\sum_{\mu}F(\sum_i\xi_{\mu i}\sigma_i)^n$. This enables the implementation of multi-spin configurations as shown by Equations \ref{eq:multi_config_1} and \ref{eq:multi_config_2}.

\begin{equation}
    E = -\sum_{\mu}F(\sum_i\xi_{\mu i}\sigma_i)^2 = -\sum_{i,j} T_{ij}\sigma_i\sigma_j
    \label{eq:multi_config_1}
\end{equation}

\begin{equation}
    E = -\sum_{\mu}F(\sum_i\xi_{\mu i}\sigma_i)^3 = -\sum_{i,j,k} T_{ijk}\sigma_i\sigma_j\sigma_k
    \label{eq:multi_config_2}
\end{equation}

where $F(x) = x^2$ where $F: \mathbb{R} \rightarrow \mathbb{R}$. Equation \ref{eq:polynomial_update_rule} is used as an update rule using this generalized energy function. It works by taking the difference of two energy functions, one with spin $i$ on and the other with it being off.

\begin{equation*}
    \sigma_{i}^{(t+1} = Sign\left[ \sum_{\mu = 1}^{K} \left( F\left( \xi_{i}^{\mu} + \sum_{j\neq i}^{N} \xi_{j}^{\mu} \sigma_{j}^{(t)} \right) - F\left( -\xi_{i}^{\mu} + \sum_{j\neq i}^{N} \xi_{j}^{\mu} \sigma_{j}^{(t)} \right) \right) \right]
    \label{eq:polynomial_update_rule}
\end{equation*}

When $n = 2$ this update rule reduces to the standard update rule. In cases with $n > 2$, each term in the energy function becomes sharper allowing a larger number of memories to be packed into the same configuration space.

Demircigil et al. \cite{exponential_Demircigil} suggest an exponential interaction function $F(x) = e^x$ that results in an energy $E = $. This exponentially increases the storage capacity of the network while still maintaining a large positive radius of attraction. This means that this increase in storage capacity does not come at the expense of associativity, and the stored patterns can still be retrieved.

Ramsauer et al. \cite{SA_update} generalize the exponential energy function by Demircigil et al. \cite{exponential_Demircigil} for continuous valued inputs while still keeping its exponential storage capacity. They propose the following new energy function: 

\begin{equation}
    E = -lse(\beta, X^T\xi) + \frac{1}{2}\xi^T\xi + \beta^{-1}logN + \frac{1}{2}M^2
    \label{eq:continous_energy}
\end{equation}

Immune repertoire classification \cite{DeepRC} uses a version of the self-attention update rule with static memory. The goal of immune repertoire classification is to find a certain sequence of immune repertoire receptors, this sequence is learned and stored in the static memory. They use a fixed query vector represented by $\xi$ as their static memory. The transformer attention mechanism is implemented as follows:

\begin{equation}
    z = softmax \left(\frac{\xi^TK^T}{\sqrt{d_k}}\right)Z
    \label{eq:DeepRC_update}
\end{equation}

Where $Z \in \mathbb{R}^{N\times d_v}$ is the value and $K \in \mathbb{R}^{N\times d_k}$ is the key.

\section{Proposed Methodology} 
\subsection{Overview}
We propose an associative memory mechanism with static memory that learns the general structure of the heart. We use the Universal Hopfield Networks (UHN) \cite{universal_networks} model to develop our mechanism. It provides a comprehensive analysis of Hopfield networks, decomposing them into three fundamental components: similarity, separation, and projection as represented by Equation \ref{eq:universal_separation}, where $P$, $sep$ and $sim$ represent projection, separation, and similarity, respectively. By separating the update rule into these three components, the UHN framework provides a more general and flexible approach to understanding and designing associative memory models.

\begin{equation}
    z = P \cdot sep (sim(M, q))
    \label{eq:universal_separation}
\end{equation}

\subsection{Main Architecture}
When a sample with a partially visible heart structure is presented to the memory module, it completes the structure and provides the model with information on the areas that are not clearly visible in the input. Our mechanism works on three matrices \textbf{K} (key), \textbf{Q} (query), and \textbf{V} (value).


\begin{figure*}[!ht]
\centering
\includegraphics[width=1\textwidth]{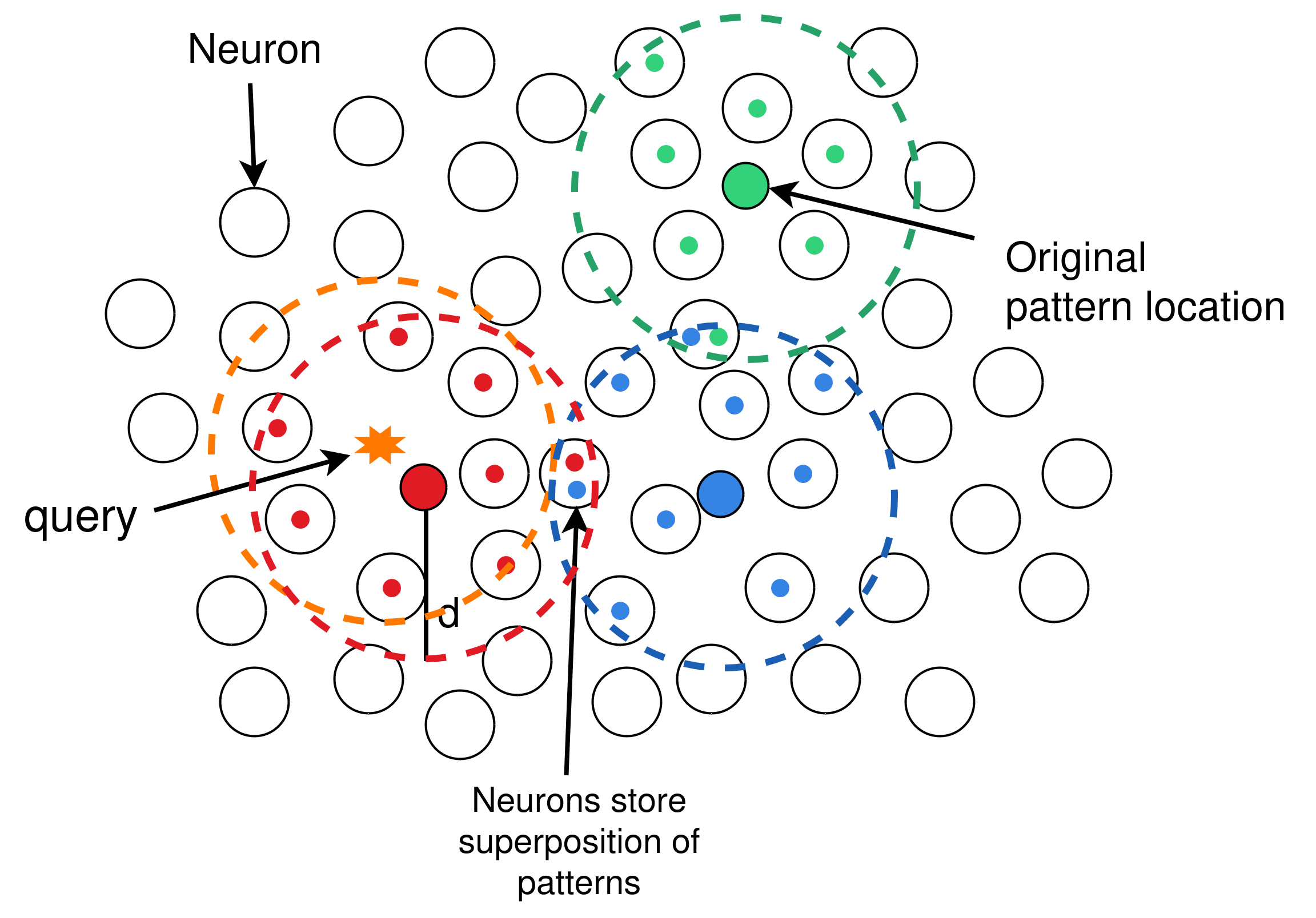}
\caption{A neuron view showing how query interacts with the stored patterns.}
\label{fig:neuron_view}
\end{figure*}

\begin{figure*}[!ht]
\centering
\includegraphics[width=1\textwidth]{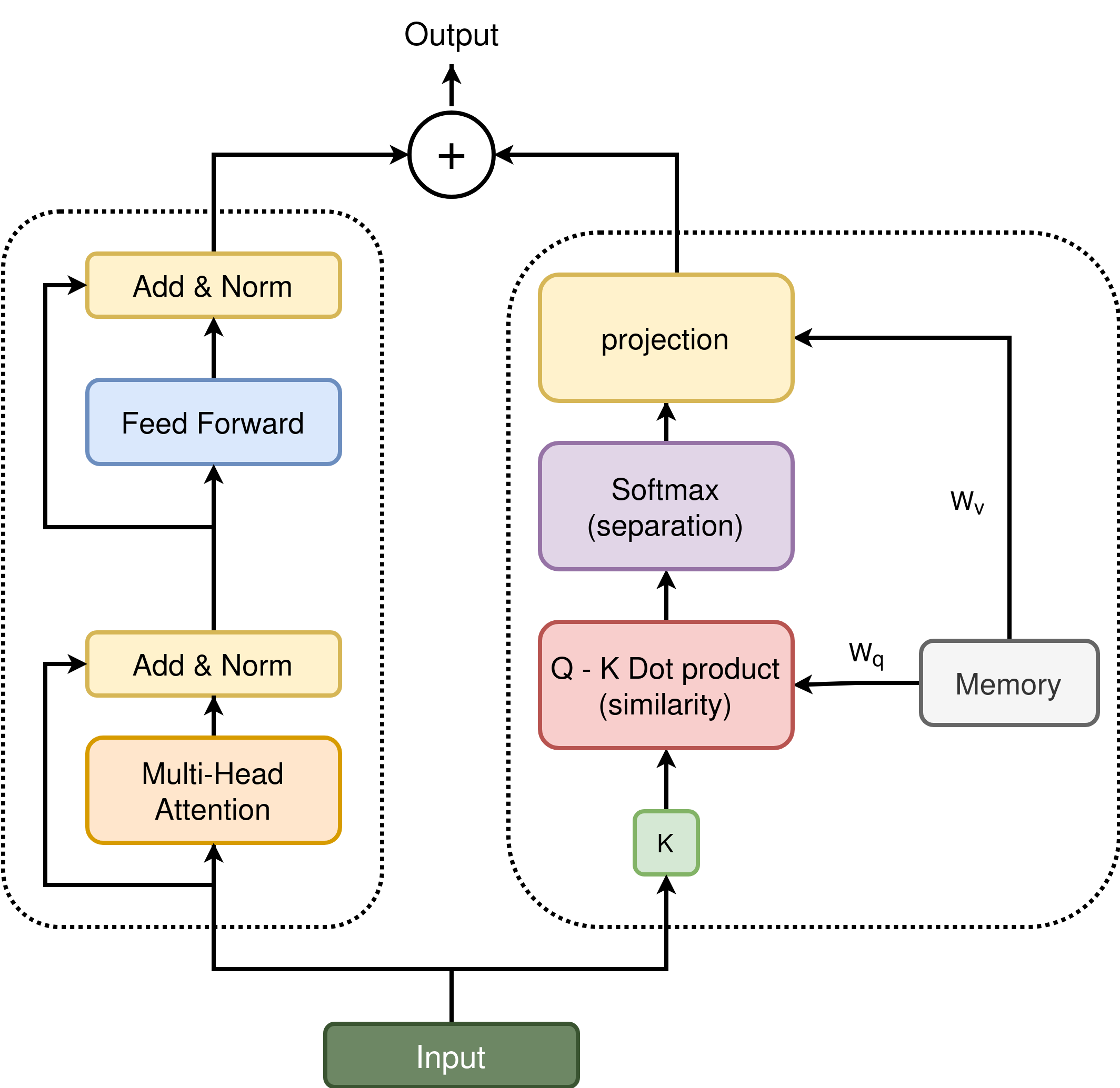}
\caption{Architectural diagram of the modified attention mechanism.}
\label{fig:architecture_diagram}
\end{figure*}

We introduce a static memory with state $\xi \in \mathbb{R}^d$. Using this static memory we set the matrices $V = W_q \cdot \xi$ and $K = W_k \cdot \xi$. Where $V \in \mathbb{R}^{m\times d}$, $K \in \mathbb{R}^{m \times d}$ and $W_q = (W_K)^T$. In our case, we use Value as the projection, softmax as separation, and dot product and similarity, resulting in an update rule shown in Equation \ref{eq:our_update}. The goal of our approach is to encourage the formation of $m$ main attractors.

\begin{equation}
    z = V \cdot Softmax \left(\frac{Q^T \cdot K}{\sqrt{d}}\right)
    \label{eq:our_update}
\end{equation}

During the learning process, the continuum of states will be clustered around these $m$ main attractors establishing $m$ meta-stable states. Each pattern is similar to every other pattern in the cardiac dataset with slight differences like orientation or size, simply training an associative memory module on all patterns would overwhelm the network and cause it to converge to a mixture of all patterns and would defeat the purpose of our memory module. Since the aim of our approach is not to learn each input pattern but to memorize the general structure of the heart in the input in multiple slight variations, it is essential to establish a small number of meta-stable states each of which contains the structure of the heart with slightly different structures.

Figure \ref{fig:neuron_view} shows a neuron view for our memory module with three meta-stable states, it demonstrates how query interacts with stored patterns. Each circular region represents an n-sphere and $d$ represents the radius of attraction for each region. The similarity between the query and the key (derived from the stored pattern) measures the cosine similarity between the query and each stored pattern. The pattern with the highest similarity is selected which in our case is the red one.

The similarity between Q and K identifies which meta-stable state is the input closely related to, and softmax as our separation function provides probabilistic interpretations of the calculated similarity. These similarity probabilities are projected onto the value vector resulting in a weighted sum of $m$ metastable states. Our update rule and energy function 
closely resembles the update rule by Model B proposed by Krotov and Hopfield \cite{largeassociativememoryproblem} where $N_F$ represents the number of inputs.


We use a transformer-based segmentation architecture called DPT \cite{dpt} as our base model. Each transformer block in this network has its corresponding memory block, the retrieved patterns from the Hopfield block are then added with the output of its respective transformer block as shown by Figure \ref{fig:architecture_diagram}. This mechanism allows the network to dynamically identify and retrieve necessary information from the memory unit and combine this information with the output of the transformer block, this addition process would fill the gaps of missing or incorrect information in the transformer output that were caused by incomplete or poorly visible patterns in the input samples.

\begin{figure*}[!ht]
\centering
\includegraphics[width=100mm]{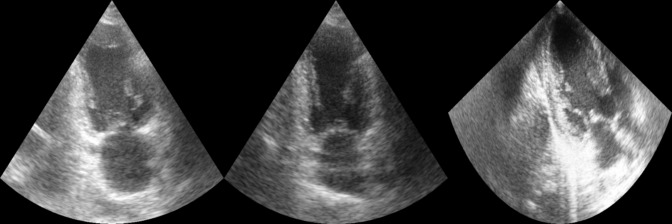}
\caption{Sample inputs from CAMUS dataset.}
\label{fig:dataset_examples}
\end{figure*}

\begin{figure*}[!ht]
\centering
\includegraphics[width=100mm]{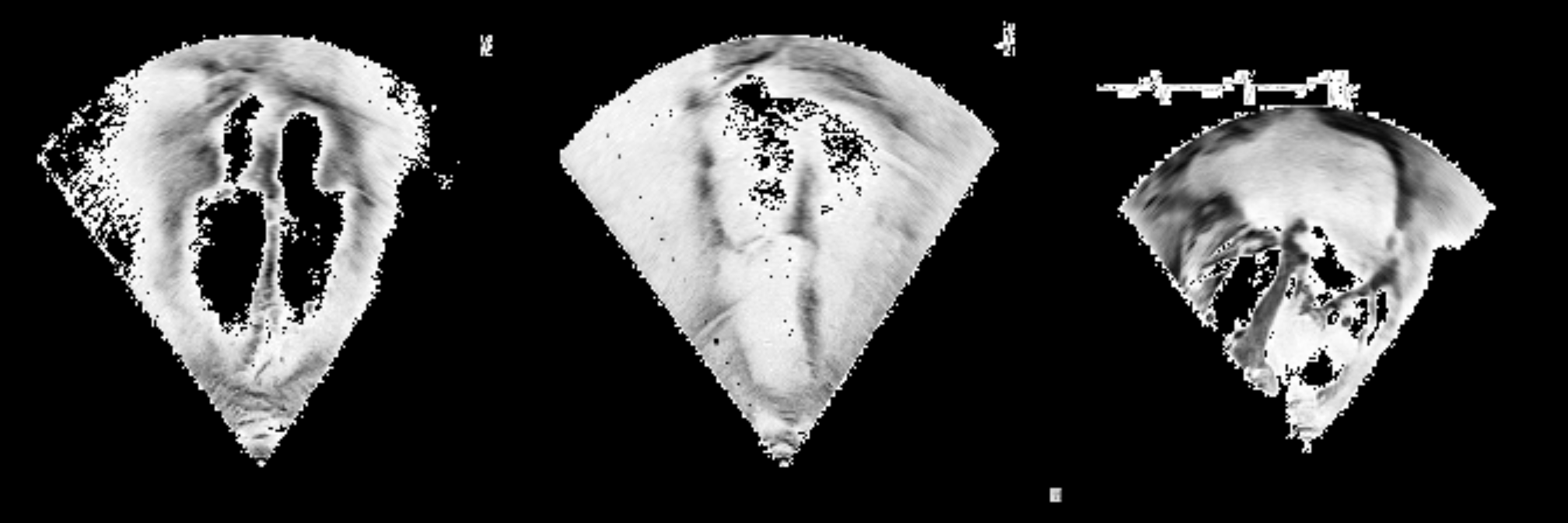}
\caption{Sample inputs from CardiacNet dataset.}
\label{fig:cardiacnet_examples}
\end{figure*}

\section{Experimental setup}
\label{experimental}
We implemented our mechanism on multiple transformer-based image segmentation mechanisms. Details of our experimental setup are explained in their respective subsections.

\subsection{Datasets}
We used Cardiac Acquisitions for Multi-structure Ultrasound Segmentation (CAMUS) \cite{camus} and CardiacNet \cite{cardiacnet} datasets for our experimentation. The CAMUS is a public dataset designed for echocardiographic image segmentation and volume estimation from 2D ultrasound sequences. It contains 2D four-chamber and two-chamber view sequences acquired from 500 patients. The entire dataset consists of a wide variability of acquisition settings and pathological cases, reflecting the slight diversity that usually occurs in real-world clinical data. We have divided the dataset so that the training dataset contains a total of 450 patients while the testing dataset contains 50 patients. Figure \ref{fig:dataset_examples} shows examples of ultrasound samples present in the CAMUS dataset.

Whereas, the CardiacNet dataset contains two sub-datasets CardiacNet-PAH and CardiacNet-ASD. CardiacNet-PAH focuses on Pulmonary Arterial Hypertension (PAH) and CardiacNet-ASD focuses on Atrial Septal Defect (ASD). Figure \ref{fig:cardiacnet_examples} shows examples of samples in the CardiacNet dataset.

\subsection{Implementation details}
All of our experiments are implemented on an RTX 3090. We used Adam optimizer with a learning rate of $5e^{-4}$ and a cosine learning rate scheduler as shown in Table \ref{table:parameter_values}. For a fair comparison, these settings were used for all of our models in our ablation study and for the training of all of the other state-of-the-art models.

\begin{table}[h!]
\centering
\scalebox{1.5}{
\scriptsize
\begin{tabular}{|c|c|}
\hline
\textbf{Parameter}   & \textbf{Value} \\ \hline
Optimizer            & Adam \\ \hline
Learning Rate        & $5 \times 10^{-4}$ \\ \hline
LR Scheduler         & Cosine \\ \hline
Warmup LR            & $1 \times 10^{-6}$ \\ \hline
Minimum LR           & $1 \times 10^{-5}$ \\ \hline
Batch Size           & 32 \\ \hline
Warmup Epochs        & 10 \\ \hline
Total Epochs         & 60 \\ \hline
Patience Epochs      & 10 \\ \hline
\end{tabular}
}
\caption{Hyperparameter values used during the training process of the model.}
\label{table:parameter_values}
\end{table}

\begin{table*}[h!]
\centering
\resizebox{\textwidth}{!}{
\scriptsize
\begin{tabular}{|c|c|c|}
\hline
\textbf{Model} & \textbf{Structure} & \textbf{Dice score} \\ \hline
\multirow{3}{*}{Hybrid DPT} & Endocardium      & 93.36\% \\ \cline{2-3} 
                            & Epicardium       & 86.14\% \\ \cline{2-3}
                            & Left atrium wall & 86.41\% \\ \hline
\multirow{3}{*}{Our Associative Hybrid DPT (DAM-Seg)} & Endocardium      & 93.72\% \\ \cline{2-3}
                                            & Epicardium       & 88.00\% \\ \cline{2-3}
                                            & Left atrium wall & 89.1\%  \\ \hline
\multirow{3}{*}{DPT Large}  & Endocardium      & 89.98\% \\ \cline{2-3}
                            & Epicardium       & 80.68\% \\ \cline{2-3}
                            & Left atrium wall & 79.27\% \\ \hline
\multirow{3}{*}{Our Associative DPT Large (DAM-Seg)}      & Endocardium      & 90.62\% \\ \cline{2-3}
                                                & Epicardium       & 81.16\% \\ \cline{2-3}
                                                & Left atrium wall & 77.76\% \\ \hline

\end{tabular}
}
\caption{Results on our ablation study with and without our associative memory module.}
\label{table:memory_ablation}
\end{table*}

\section{Results} \label{results}
\subsection{Ablation studies}
\begin{figure*}
\centering
\includegraphics[width=130mm]{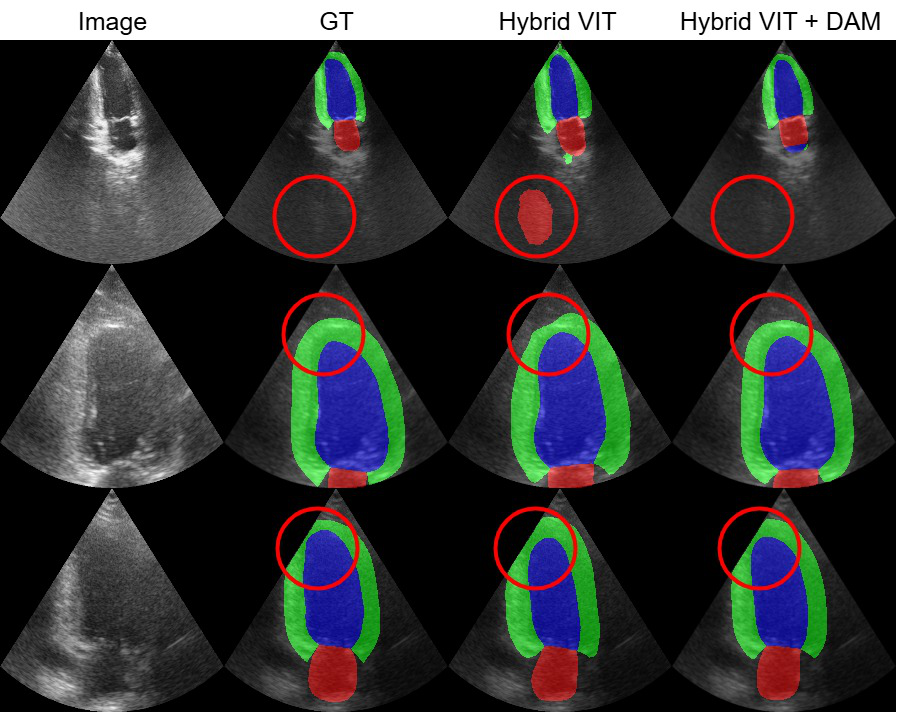}
\caption{Qualitative comparison between the Ground Truth and segmentation results for hybrid VIT and hybrid VIT + DAM, where blue, green, and red represent the Endocardium, Epicardium, and left atrium wall.}
\label{fig:visual_results}
\end{figure*}

We investigated the contributions and the effectiveness of our memory module on variations of DPT architecture. We have experimented on Hybrid DPT and DPT Large with and without our dense associative module as shown in Table \ref{table:memory_ablation}. Our associative memory module seems to affect the performance of all three classes but has the most influence over the performance of Epicardium and the left atrium wall since these are usually affected by poor quality and visibility.
In Figure \ref{fig:visual_results}, We present some examples for qualitatively comparing the segmentation quality between hybrid VIT and our proposed hybrid VIT + DAM and how the improvement of our memory module translates visually.

\subsection{Comparison with recent SOTA methods}
This section discusses the comparison between our proposed mechanism against other state-of-the-art architectures. These comparative results also show a similar trend in performance improvements as in the ablation study. Since Endocardium is least affected by poor image quality and rarely has any visibility issues the performance on Endocardium almost stays the same for each architecture as shown in Table \ref{table:sota_comparison}. Epicardium and the Left atrium wall show the largest performance improvement. We also performed quantitative comparative analysis on the CardiacNet dataset, the results in Table \ref{table:sota_comparison_card} show similar performance improvements, implying that our methodology is robust against varying scenarios.

\begin{table*}[h!]
\centering
\resizebox{\textwidth}{!}{
\begin{tabular}{|c|c|c|}
\hline
\textbf{Model} & \textbf{Structure} & \textbf{Dice score} \\ \hline
\multirow{3}{*}{UNet\cite{unet}}& Endocardium      & 93.69\% \\ 
                                & Epicardium       & 86.25\% \\ 
                                & Left atrium wall & 85.34\% \\ \hline
\multirow{3}{*}{nnUNet\cite{nnUnet}}& Endocardium      & 93.33\% \\ 
                                    & Epicardium       & 87.06\% \\ 
                                    & Left atrium wall & 85.13\%  \\ \hline
\multirow{3}{*}{CANet\cite{CANet}}& Endocardium      & 93.91\% \\ 
                                  & Epicardium       & 87.52\% \\ 
                                  & Left atrium wall & 87.01\% \\ \hline
\multirow{3}{*}{Extended nnUNet\cite{extended_nnunet}}   & Endocardium      & 92.90\% \\ 
                                                         & Epicardium       & 85.81\% \\ 
                                                         & Left atrium wall & 86.54\% \\ \hline
\multirow{3}{*}{\textbf{Our Associative Hybrid DPT (DAM-Seg)}}  & Endocardium      & \textbf{93.72}\% \\ 
                                             & Epicardium       & \textbf{88.00}\% \\ 
                                             & Left atrium wall & \textbf{89.10}\% \\ \hline

\end{tabular}
}
\caption{Comparative results against cardiac segmentation mechanisms on CAMUS dataset.}
\label{table:sota_comparison}
\end{table*}

\begin{table*}[ht!]
\centering
\resizebox{\textwidth}{!}{
\begin{tabular}{|c|c|c|}
\hline
\textbf{Model} & \textbf{Structure} & \textbf{Dice score} \\ \hline
\multirow{4}{*}{UNet\cite{unet}}& Left Atrium     & 88.61\% \\ 
                                & Right Atrium    & 87.98\% \\ 
                                & Left Ventricle  & 91.51\% \\ 
                                & Right Ventricle & 88.19\% \\ \hline
\multirow{4}{*}{nnUNet\cite{nnUnet}}  & Left Atrium     & 88.66\% \\ 
                                      & Right Atrium    & 88.47\% \\ 
                                      & Left Ventricle  & 91.40\%  \\ 
                                      & Right Ventricle & 88.50\%  \\ \hline
\multirow{4}{*}{CANet\cite{CANet}}  & Left Atrium     & 88.95\% \\ 
                                    & Right Atrium    & 88.91\% \\ 
                                    & Left Ventricle  & 91.70\% \\ 
                                    & Right Ventricle & 89.26\% \\ \hline
\multirow{4}{*}{Extended nnUNet\cite{extended_nnunet}}  & Left Atrium     & 88.21\% \\ 
                                                        & Right Atrium    & 88.23\% \\ 
                                                        & Left Ventricle  & 91.03\% \\ 
                                                        & Right Ventricle & 88.28\% \\ \hline
\multirow{4}{*}{\textbf{DAM-Seg}}  & Left Atrium     & \textbf{89.52}\% \\ 
                                             & Right Atrium    & \textbf{87.68}\% \\ 
                                             & Left Ventricle  & \textbf{91.84}\% \\ 
                                             & Right Ventricle & \textbf{89.82}\% \\ \hline

\end{tabular}
}
\caption{Comparative results against cardiac segmentation mechanisms on CardiacNet dataset.}
\label{table:sota_comparison_card}
\end{table*}


\begin{figure*}
\centering
\includegraphics[width=0.7\textwidth]{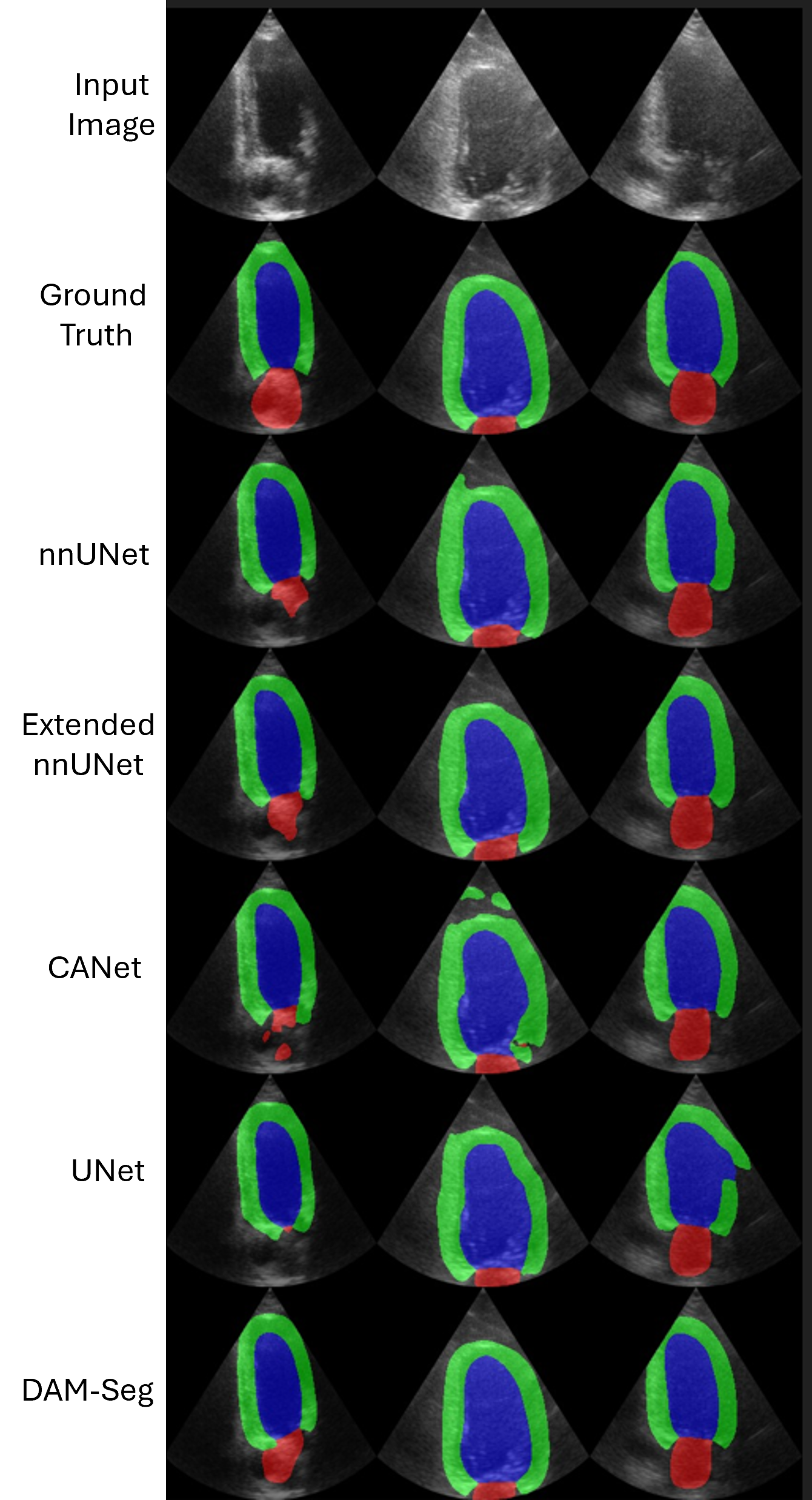}
\caption{Visual comparison between our methodology (DAM-Seg) and its counterparts, where blue, green, and red represent the Endocardium, Epicardium, and left atrium wall.}
\label{fig:comparative_visual_results}
\end{figure*}

\begin{figure*}
\centering
\scriptsize
\includegraphics[width=1\textwidth]
{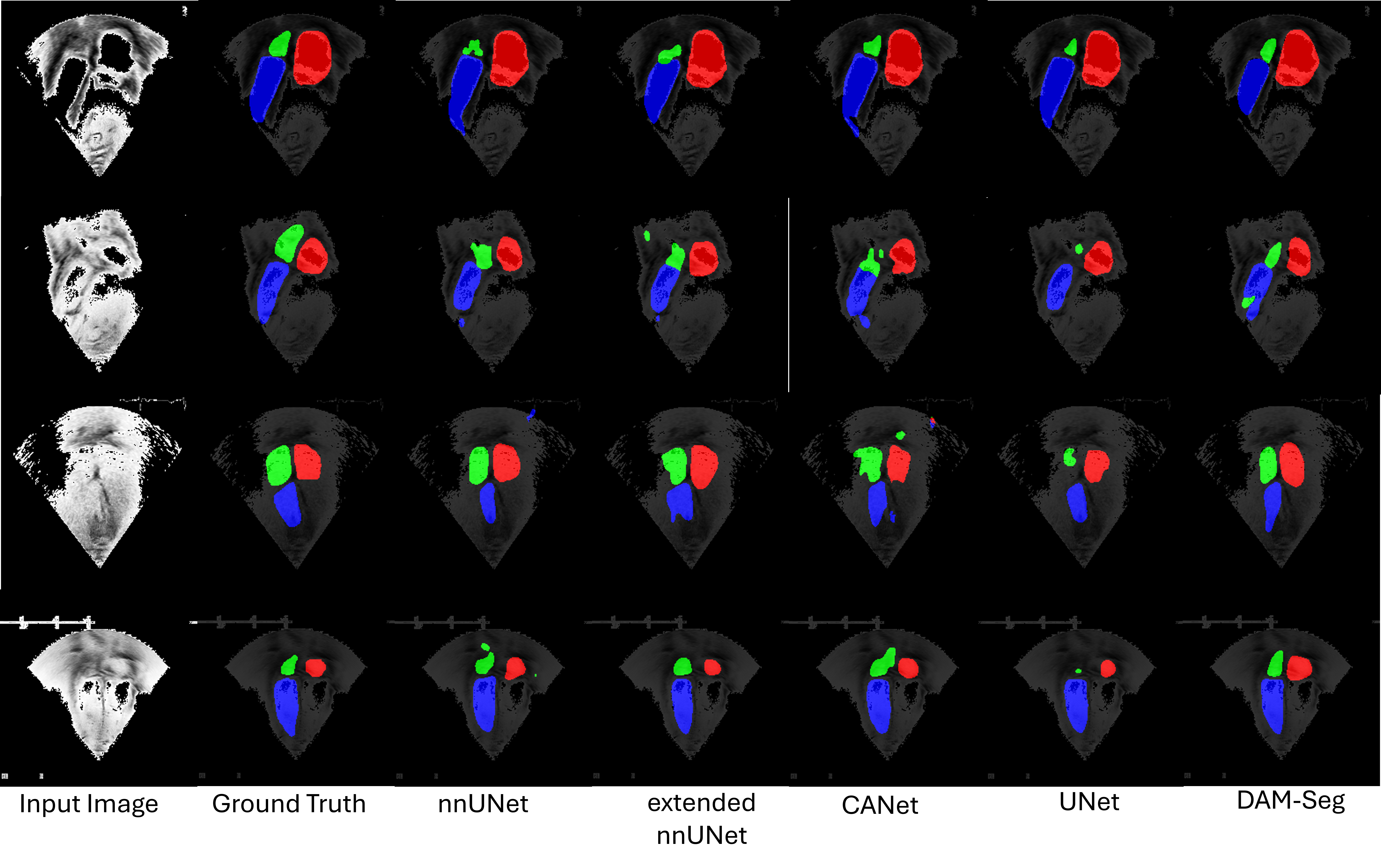}
\caption{Segmentation results of CardiacNet dataset from nnUNet, extended nnUNet, CANet, UNet, and our proposed network (DAM-Seg) has been shown. Where red, green, and blue represent Left ventricle, right ventricle, left atrium, respectively.}
\label{fig:cardiac_dataset}
\end{figure*}

\section{Discussion}
While modern machine learning techniques have demonstrated superior performance to humans in numerous classification tasks, there is a growing concern about their lack of true understanding of the underlying structure of training data as highlighted in studies such as  \cite{properties_nn, nn_easily}. A critical gap exists between machine learning models and human cognition. Deep neural networks excel at pattern recognition within their training distribution, however, they often fail to grasp the semantic and contextual understanding that humans naturally apply. Unlike deep learning algorithms we humans tend to have a general mental idea of the structure we are trying to segment or detect. This mental image helps us to easily identify the desired object even in noisy environments. This discrepancy between humans and deep neural networks raises important questions about the robustness, reliability, and interpretability of machine learning models, especially in high-stakes applications where errors could have significant consequences. Addressing these limitations is essential to ensure the safe and effective deployment of such models in critical domains.

The issue of anatomically incorrect segmentation has been extensively reported and studied by multiple papers \cite{adiga2024anatomically,van2024towards,wyburd2024anatomically,gao2023anatomy,von2023anatomically}, this problem often stems from the neural network's inability to learn and store a mental image of the object of interest. Research efforts, such as those in \cite{VAE_wrapping, GCN} have proposed methods to bypass the need for the network to learn a mental image of patterns by proposing other ways of maintaining anatomical accuracy.
The ability of dense associative memories to store and retrieve complex patterns efficiently aligns closely with human cognitive processes, particularly in the context of visual perception and segmentation tasks. Inspired by this, we propose a dense associative memory module enabling neural networks to have a "mental image" of objects they are trying to segment, this capability helps the network to preserve anatomical structure in its segmentation results, even on images with poor visibility.

\subsection{Visual analysis}

To further expand the comparison, we also visualize the results of our model i.e., DAM-seg, against its other counterparts. Figure \ref{fig:comparative_visual_results} and Figure \ref{fig:cardiac_dataset} show visual results on CAMUS and CardiacNet datasets, respectively, where we compare the segmentation results with corresponding original images, ground truths, nnUNet, extended nnUNet, CANet, UNet, and our proposed DAM-segNet. Figure \ref{fig:comparative_visual_results} shows that our model can perform well even for instances where the pattern is not clearly visible while other models tend to struggle to establish accurate boundaries, this is prevalent mainly for the left atrium wall. The CardiacNet dataset does not suffer from poor patterns in input images as much as the CAMUS dataset does, therefore, the results seem similar for all models in Figure \ref{fig:cardiac_dataset} except few minor details. Thus, we can conclude that our proposed model is effective for cardiac
segmentation tasks. It can also be noted that the performance of the proposed scheme remains consistent for the left ventricle, right ventricle, and left atrium.


\subsection{Limitation and future works}
While our mechanism has improved our segmentation results, it comes with multiple drawbacks.
Firstly, the reliance on the fixed memory tensor may limit the model's ability to adapt to diverse or evolving input distributions, potentially reducing its generalization capabilities. The static nature of the keys and values might lead to a less dynamic representation of the input space, possibly resulting in reduced expressiveness for context-dependent tasks. It could still be very effective for tasks where prior knowledge is crucial, but less so for tasks requiring high input-specific adaptability.
In the future we may address these issues by implementing a forgetting mechanism similar to the one described in the Memformer model \cite{memformer}. This would help in managing the stored information over time, preventing it from becoming stale or overly dominant. A Memory Retention Valve (MRV) mentioned in the RATE model \cite{rate} to control how much of the old information in A is retained versus how much new information is incorporated can also be a potential alternative.

\section{Conclusions} \label{con}
This paper presents a novel approach to enhancing static memory within attention mechanisms. A fixed memory tensor is utilized to generate keys and values through learned projection matrices, while queries remain dynamically dependent on the input. Experimental results demonstrate that incorporating this memory module into a transformer-based segmentation model significantly improves performance. The stored memory effectively aids the model in producing anatomically accurate segmentation masks, even in scenarios where input structures are partially obscured or poorly visible. This capability highlights the potential of the proposed method for robust and reliable segmentation when dealing with inputs with partially improperly visible structures.

\section*{CRediT authorship contribution statement}
\textbf{Zahid Ullah:} Conceptualization, Methodology, Software, Formal analysis, Investigation, Writing - original draft, Writing - review \& editing.  \textbf{Jihie Kim:} Formal analysis, Investigation, Supervision, Project administration.  

\section*{\textbf{Declaration of Competing Interests}} The authors declare that they have no known competing financial interests or personal relationships that could have appeared to influence the work reported in this paper.

\section{Data and code availability statement}
The datasets used in this research are publicly accessible through the OpenBHB Challenge. The code, along with the trained model weights, can be obtained by contacting the corresponding author via email.

\section*{Acknowledgements}
This research was supported by the MSIT(Ministry of Science and ICT), Korea, under the ITRC(Information Technology Research Center) support program(IITP-2025-RS-2020-II201789), and the Artificial Intelligence Convergence Innovation Human Resources Development(IITP-2025-RS-2023-00254592) supervised by the IITP(Institute for Information \& Communications Technology Planning \& Evaluation).

\bibliographystyle{IEEEtran}
\bibliography{Reference}

@article{adiga2024anatomically,
  title={Anatomically-aware uncertainty for semi-supervised image segmentation},
  author={Adiga, Sukesh and Dolz, Jose and Lombaert, Herve},
  journal={Medical Image Analysis},
  volume={91},
  pages={103011},
  year={2024},
  publisher={Elsevier}
}

@article{van2024towards,
  title={Towards Robust Cardiac Segmentation using Graph Convolutional Networks},
  author={Van De Vyver, Gilles and Thomas, Sarina and Ben-Yosef, Guy and Olaisen, Sindre Hellum and Dalen, H{\aa}vard and L{\o}vstakken, Lasse and Smistad, Erik},
  journal={IEEE Access},
  year={2024},
  publisher={IEEE}
}

@article{wyburd2024anatomically,
  title={Anatomically plausible segmentations: Explicitly preserving topology through prior deformations},
  author={Wyburd, Madeleine K and Dinsdale, Nicola K and Jenkinson, Mark and Namburete, Ana IL},
  journal={Medical Image Analysis},
  pages={103222},
  year={2024},
  publisher={Elsevier}
}

@article{gao2023anatomy,
  title={An anatomy-aware framework for automatic segmentation of parotid tumor from multimodal MRI},
  author={Gao, Yifan and Dai, Yin and Liu, Fayu and Chen, Weibing and Shi, Lifu},
  journal={Computers in Biology and Medicine},
  volume={161},
  pages={107000},
  year={2023},
  publisher={Elsevier}
}

@article{von2023anatomically,
  title={Anatomically-guided deep learning for left ventricle geometry generation with uncertainty quantification based on short-axis MR images},
  author={Von Zuben, Andre and Perotti, Luigi E and Viana, Felipe AC},
  journal={Engineering Applications of Artificial Intelligence},
  volume={121},
  pages={106012},
  year={2023},
  publisher={Elsevier}
}

@misc{polynomial_DAM,
      title={Dense Associative Memory for Pattern Recognition}, 
      author={Dmitry Krotov and John J Hopfield},
      year={2016},
      eprint={1606.01164},
      archivePrefix={arXiv},
      primaryClass={cs.NE},
      url={https://arxiv.org/abs/1606.01164}, 
}

@article{exponential_Demircigil,
   title={On a Model of Associative Memory with Huge Storage Capacity},
   volume={168},
   ISSN={1572-9613},
   url={http://dx.doi.org/10.1007/s10955-017-1806-y},
   DOI={10.1007/s10955-017-1806-y},
   number={2},
   journal={Journal of Statistical Physics},
   publisher={Springer Science and Business Media LLC},
   author={Demircigil, Mete and Heusel, Judith and Löwe, Matthias and Upgang, Sven and Vermet, Franck},
   year={2017},
   month=may, pages={288–299} }

@misc{SA_update,
      title={Hopfield Networks is All You Need}, 
      author={Hubert Ramsauer and Bernhard Schäfl and Johannes Lehner and Philipp Seidl and Michael Widrich and Thomas Adler and Lukas Gruber and Markus Holzleitner and Milena Pavlović and Geir Kjetil Sandve and Victor Greiff and David Kreil and Michael Kopp and Günter Klambauer and Johannes Brandstetter and Sepp Hochreiter},
      year={2021},
      eprint={2008.02217},
      archivePrefix={arXiv},
      primaryClass={cs.NE},
      url={https://arxiv.org/abs/2008.02217}, 
}

@misc{DeepRC,
      title={Modern Hopfield Networks and Attention for Immune Repertoire Classification}, 
      author={Michael Widrich and Bernhard Schäfl and Hubert Ramsauer and Milena Pavlović and Lukas Gruber and Markus Holzleitner and Johannes Brandstetter and Geir Kjetil Sandve and Victor Greiff and Sepp Hochreiter and Günter Klambauer},
      year={2020},
      eprint={2007.13505},
      archivePrefix={arXiv},
      primaryClass={cs.LG},
      url={https://arxiv.org/abs/2007.13505}, 
}

@misc{universal_networks,
      title={Universal Hopfield Networks: A General Framework for Single-Shot Associative Memory Models}, 
      author={Beren Millidge and Tommaso Salvatori and Yuhang Song and Thomas Lukasiewicz and Rafal Bogacz},
      year={2022},
      eprint={2202.04557},
      archivePrefix={arXiv},
      primaryClass={cs.NE},
      url={https://arxiv.org/abs/2202.04557}, 
}

@misc{dpt,
      title={Vision Transformers for Dense Prediction}, 
      author={René Ranftl and Alexey Bochkovskiy and Vladlen Koltun},
      year={2021},
      eprint={2103.13413},
      archivePrefix={arXiv},
      primaryClass={cs.CV},
      url={https://arxiv.org/abs/2103.13413}, 
}

@article{RV_segmentation,
  title={Right ventricle segmentation from cardiac MRI: a collation study},
  author={Petitjean, Caroline and Zuluaga, Maria A and Bai, Wenjia and Dacher, Jean-Nicolas and Grosgeorge, Damien and Caudron, J{\'e}r{\^o}me and Ruan, Su and Ayed, Ismail Ben and Cardoso, M Jorge and Chen, Hsiang-Chou and others},
  journal={Medical image analysis},
  volume={19},
  number={1},
  pages={187--202},
  year={2015},
  publisher={Elsevier}
}

@misc{fcn,
      title={Fully Convolutional Networks for Semantic Segmentation}, 
      author={Jonathan Long and Evan Shelhamer and Trevor Darrell},
      year={2015},
      eprint={1411.4038},
      archivePrefix={arXiv},
      primaryClass={cs.CV},
      url={https://arxiv.org/abs/1411.4038}, 
}

@misc{cardiac_fcn,
      title={A Fully Convolutional Neural Network for Cardiac Segmentation in Short-Axis MRI}, 
      author={Phi Vu Tran},
      year={2017},
      eprint={1604.00494},
      archivePrefix={arXiv},
      primaryClass={cs.CV},
      url={https://arxiv.org/abs/1604.00494}, 
}

@ARTICLE{VAE_wrapping,
  author={Painchaud, Nathan and Skandarani, Youssef and Judge, Thierry and Bernard, Olivier and Lalande, Alain and Jodoin, Pierre-Marc},
  journal={IEEE Transactions on Medical Imaging}, 
  title={Cardiac Segmentation With Strong Anatomical Guarantees}, 
  year={2020},
  volume={39},
  number={11},
  pages={3703-3713},
  doi={10.1109/TMI.2020.3003240}}

@ARTICLE{GCN,
  author={Vyver, Gilles Van De and Thomas, Sarina and Ben-Yosef, Guy and Olaisen, Sindre Hellum and Dalen, Håvard and Løvstakken, Lasse and Smistad, Erik},
  journal={IEEE Access}, 
  title={Toward Robust Cardiac Segmentation Using Graph Convolutional Networks}, 
  year={2024},
  volume={12},
  number={},
  pages={33876-33888},
  doi={10.1109/ACCESS.2024.3373046}}

@article{hopfield_orig,
  title={Neural networks and physical systems with emergent collective computational abilities.},
  author={Hopfield, John J},
  journal={Proceedings of the national academy of sciences},
  volume={79},
  number={8},
  pages={2554--2558},
  year={1982},
  publisher={National Acad Sciences}
}

@misc{properties_nn,
      title={Intriguing properties of neural networks}, 
      author={Christian Szegedy and Wojciech Zaremba and Ilya Sutskever and Joan Bruna and Dumitru Erhan and Ian Goodfellow and Rob Fergus},
      year={2014},
      eprint={1312.6199},
      archivePrefix={arXiv},
      primaryClass={cs.CV},
      url={https://arxiv.org/abs/1312.6199}, 
}

@misc{nn_easily,
      title={Deep Neural Networks are Easily Fooled: High Confidence Predictions for Unrecognizable Images}, 
      author={Anh Nguyen and Jason Yosinski and Jeff Clune},
      year={2015},
      eprint={1412.1897},
      archivePrefix={arXiv},
      primaryClass={cs.CV},
      url={https://arxiv.org/abs/1412.1897}, 
}

@article{camus,
  title={Deep Learning for Segmentation Using an Open Large-Scale Dataset in 2D Echocardiography},
  author={Sarah Leclerc and Erik Smistad and Jo{\~a}o Pedrosa and Andreas {\O}stvik and Fr{\'e}d{\'e}ric Cervenansky and Florian Espinosa and Torvald Espeland and Erik Andreas Rye Berg and Pierre-Marc Jodoin and Thomas Grenier and Carole Lartizien and Jan D’hooge and Lasse L{\o}vstakken and Olivier Bernard},
  journal={IEEE Transactions on Medical Imaging},
  year={2019},
  volume={38},
  pages={2198-2210},
  url={https://api.semanticscholar.org/CorpusID:73510235}
}

@ARTICLE{displacement_aware,
  author={Li, Kang and Zhu, Yu and Yu, Lequan and Heng, Pheng-Ann},
  journal={IEEE Transactions on Medical Imaging}, 
  title={A Dual Enrichment Synergistic Strategy to Handle Data Heterogeneity for Domain Incremental Cardiac Segmentation}, 
  year={2024},
  volume={43},
  number={6},
  pages={2279-2290},
  doi={10.1109/TMI.2024.3364240}}

@INPROCEEDINGS{cross_domain,
  author={Cai, Zhuotong and Xin, Jingmin and Dong, Siyuan and Onofrey, John A. and Zheng, Nanning and Duncan, James S.},
  booktitle={ICASSP 2024 - 2024 IEEE International Conference on Acoustics, Speech and Signal Processing (ICASSP)}, 
  title={Symmetric Consistency with Cross-Domain Mixup for Cross-Modality Cardiac Segmentation}, 
  year={2024},
  volume={},
  number={},
  pages={1536-1540},
  doi={10.1109/ICASSP48485.2024.10447304}}

@misc{memformer,
      title={Memformer: A Memory-Augmented Transformer for Sequence Modeling}, 
      author={Qingyang Wu and Zhenzhong Lan and Kun Qian and Jing Gu and Alborz Geramifard and Zhou Yu},
      year={2022},
      eprint={2010.06891},
      archivePrefix={arXiv},
      primaryClass={cs.CL},
      url={https://arxiv.org/abs/2010.06891}, 
}

@misc{rate,
      title={Recurrent Action Transformer with Memory}, 
      author={Egor Cherepanov and Alexey Staroverov and Dmitry Yudin and Alexey K. Kovalev and Aleksandr I. Panov},
      year={2024},
      eprint={2306.09459},
      archivePrefix={arXiv},
      primaryClass={cs.LG},
      url={https://arxiv.org/abs/2306.09459}, 
}

@article{CANet,
   title={CANet: An Unsupervised Intrusion Detection System for High Dimensional CAN Bus Data},
   volume={8},
   ISSN={2169-3536},
   url={http://dx.doi.org/10.1109/ACCESS.2020.2982544},
   DOI={10.1109/access.2020.2982544},
   journal={IEEE Access},
   publisher={Institute of Electrical and Electronics Engineers (IEEE)},
   author={Hanselmann, Markus and Strauss, Thilo and Dormann, Katharina and Ulmer, Holger},
   year={2020},
   pages={58194–58205} }

@misc{nnUnet,
      title={nnU-Net: Self-adapting Framework for U-Net-Based Medical Image Segmentation}, 
      author={Fabian Isensee and Jens Petersen and Andre Klein and David Zimmerer and Paul F. Jaeger and Simon Kohl and Jakob Wasserthal and Gregor Koehler and Tobias Norajitra and Sebastian Wirkert and Klaus H. Maier-Hein},
      year={2018},
      eprint={1809.10486},
      archivePrefix={arXiv},
      primaryClass={cs.CV},
      url={https://arxiv.org/abs/1809.10486}, 
}

@misc{extended_nnunet,
      title={Extending nnU-Net is all you need}, 
      author={Fabian Isensee and Constantin Ulrich and Tassilo Wald and Klaus H. Maier-Hein},
      year={2022},
      eprint={2208.10791},
      archivePrefix={arXiv},
      primaryClass={eess.IV},
      url={https://arxiv.org/abs/2208.10791}, 
}

@misc{unet,
      title={U-Net: Convolutional Networks for Biomedical Image Segmentation}, 
      author={Olaf Ronneberger and Philipp Fischer and Thomas Brox},
      year={2015},
      eprint={1505.04597},
      archivePrefix={arXiv},
      primaryClass={cs.CV},
      url={https://arxiv.org/abs/1505.04597}, 
}

@misc{cardiacnet,
      title={GraphEcho: Graph-Driven Unsupervised Domain Adaptation for Echocardiogram Video Segmentation}, 
      author={Jiewen Yang and Xinpeng Ding and Ziyang Zheng and Xiaowei Xu and Xiaomeng Li},
      year={2023},
      eprint={2309.11145},
      archivePrefix={arXiv},
      primaryClass={cs.CV},
      url={https://arxiv.org/abs/2309.11145}, 
}

@misc{largeassociativememoryproblem,
      title={Large Associative Memory Problem in Neurobiology and Machine Learning}, 
      author={Dmitry Krotov and John Hopfield},
      year={2021},
      eprint={2008.06996},
      archivePrefix={arXiv},
      primaryClass={q-bio.NC},
      url={https://arxiv.org/abs/2008.06996}, 
}

\end{document}